\pdfoutput=1

\documentclass[11pt]{article}

\usepackage[]{eacl2023}

\usepackage{times}
\usepackage{latexsym}
\usepackage{nicematrix}
\usepackage{amsmath}
\usepackage{amssymb}

\usepackage[T1]{fontenc}

\usepackage[utf8]{inputenc}
\usepackage{svg}
\usepackage{subcaption}
\usepackage{xspace}
\usepackage{tabularx}
\usepackage{euflag}

\usepackage{microtype}
\usepackage{inconsolata}
\usepackage{booktabs}
\newcommand{\mmmpd}[1]{M3P$^{D}$\xspace}
\newcommand{\mmmp}[1]{M3P\xspace}
\newcommand{\mdoq}[1]{M3P$^{Q}$\xspace}
\newcommand{\msboot}[1]{M3P$^{Q}$ + SB\xspace}

\newcommand{\uccd}[1]{UC2$^{D}$\xspace}
\newcommand{\ucc}[1]{UC2\xspace}
\newcommand{\ucdoq}[1]{UC2$^{Q}$\xspace}
\newcommand{\ucsboot}[1]{UC2$^{Q}$ + SB\xspace}

\newcommand{\sparagraph}[1]{\noindent\textbf{#1.}}
\newcommand{\rparagraph}[1]{\vspace{1.6mm}\noindent\textbf{#1.}}
\newcolumntype{Y}{>{\centering\arraybackslash}X}

\usepackage{xcolor, colortbl}

\definecolor{nice-red}{HTML}{E41A1C}
\definecolor{nice-orange}{HTML}{FF7F00}
\definecolor{nice-yellow}{HTML}{FFC020}
\definecolor{nice-green}{HTML}{4DAF4A}
\definecolor{nice-blue}{HTML}{377EB8}
\definecolor{nice-purple}{HTML}{984EA3}
\definecolor{gray-blue}{HTML}{92c5de}
\definecolor{gray-red}{HTML}{edb7bd}

\newcommand{\gbc}{\cellcolor{gray-blue}}

\usepackage{todonotes}
\makeatletter
\newcommand*\iftodonotes{\if@todonotes@disabled\expandafter\@secondoftwo\else\expandafter\@firstoftwo\fi}
\makeatother

\title{Delving Deeper into Cross-lingual Visual Question Answering}

\author{\bf Chen Cecilia Liu$^1$, Jonas Pfeiffer$^{1}$, Anna Korhonen$^{2}$,
{\bf Ivan Vuli\'{c}$^{2}$, Iryna Gurevych$^{1}$ } \\
$^1$Ubiquitous Knowledge Processing Lab,\\ Department of Computer Science and Hessian Center for AI (hessian.AI), \\ Technical University of Darmstadt \\  
$^2$Language Technology Lab, University of Cambridge \hspace{0.5em} \\
{\url{www.ukp.tu-darmstadt.de}} \\
}

\begin{document}
\maketitle
\begin{abstract}
Visual question answering (VQA) is one of the crucial vision-and-language tasks. Yet, existing VQA research has mostly focused on the English language, due to a lack of suitable evaluation resources. Previous work on cross-lingual VQA has reported poor zero-shot transfer performance of current multilingual multimodal Transformers with large gaps to monolingual performance, without any deeper analysis. In this work, we delve deeper into the different aspects of cross-lingual VQA, aiming to understand the impact of 1) modeling methods and choices, including architecture, inductive bias, fine-tuning; 2) learning biases: including question types and modality biases in cross-lingual setups. The key results of our analysis are: \textbf{1)}~We show that simple modifications to the standard training setup can substantially reduce the transfer gap to monolingual English performance, yielding +10 accuracy points over existing methods. \textbf{2)}~We analyze cross-lingual VQA across different question types of varying complexity for different multilingual multimodal Transformers, and identify question types that are the most difficult to improve on. \textbf{3)}~We provide an analysis of modality biases present in training data and models, revealing why zero-shot performance gaps remain for certain question types and languages. 

\end{abstract}

\section{Introduction}
\label{s:intro}

The lack of multilingual resources has hindered the development and evaluation of Visual Question Answering (VQA) methods beyond the English language until recently. A rise in interest in creating multilingual Vision-and-Language (V\&L) resources has inspired more research in this area \cite[\textit{inter alia}]{Srinivasan2021WIT,su-etal-2021-gem,liu-etal-2021-visually,pfeiffer2021xgqa,Wang2021MultiSubs, bugliarello2022IGLUE}. Large Transformer-based models pretrained on images and text in \textit{multiple} different languages have been proven as a viable vehicle for the development of multilingual V\&L task architectures through transfer learning, but such models are still few and far between \cite[M3P, UC2;][]{huang2020m3p,zhou2021uc}. Large decreases in task performance between monolingual and (zero-shot) cross-lingual transfer setups have been measured and reported, among other multilingual V\&L tasks, in VQA \cite{pfeiffer2021xgqa}. Yet, the reasons for such low results in this pivotal V\&L task have not been investigated in depth. 

In this work, we aim to shed new light on the cross-lingual performance gap of cross-lingual VQA models from multiple angles. To the best of our knowledge, we are the first to provide a comprehensive analysis of multilingual VQA, with a focus on cross-lingual transfer. 

We first assess and discuss the impact of modeling methods and choices on the final cross-lingual VQA performance, aiming to mitigate the present performance gap. This includes experimenting with diverse prediction head architectures, incorporating inductive bias by extending input signals, as well as more sophisticated fine-tuning strategies. We analyze cross-lingual VQA across different question types of varying complexity for different multilingual multimodal Transformers, and in zero-shot and few-shot scenarios. 

Next, we focus on the learning biases, where we investigate whether current multilingual multimodal models suffer from the so-called unimodal bias: that is, we probe if the models truly reason over both images and questions to solve the VQA task, or if they take unimodal `shortcuts' instead, exploiting spurious correlations and artifacts of data creation. Our analysis allows us to identify the most difficult question types and reveals a shortcoming of the current evaluation scheme.

We find that standard approaches from text-only cross-lingual transfer scenarios \cite{pires-etal-2019-multilingual,xtreme} do not leverage the full multilingual capabilities of the pretrained models; we measure the considerably worse performance of `standard' fine-tuning compared to a simple modified fine-tuning regime. Interestingly, we report a discrepancy between monolingual and cross-lingual performance in the modified fine-tuning regime: while they do not have any substantial impact on the model performance in the \textit{source} language (English), they considerably improve \textit{cross-lingual} VQA capabilities, achieving gains of more than 10 absolute accuracy points over the baselines.

Code is available at \url{github.com/UKPLab/eacl2023-xlingvqa}. 

\section{Preliminaries}
\label{s:background}

The VQA task is typically framed as a classification problem with a large number of classes. For instance, in the VQA task on the standard English GQA dataset~\cite{hudson2018gqa}, given a pair of an image and a question, a model needs to predict a correct answer from 1,853 possible classes. GQA consists of diverse structural and semantic patterns, in which the questions are visually grounded in the image. In multilingual and cross-lingual VQA, the goal is to make similar predictions, but the questions can be posed in different \textit{target} languages \cite{pfeiffer2021xgqa}: e.g., the VQA task on the multilingual xGQA dataset~\cite{pfeiffer2021xgqa} relies on the same set of 1,853 classes as English GQA.

We base all our analyses and experiments on the xGQA dataset, which is, due to its size and language coverage, arguably the most comprehensive evaluation resource for cross-lingual VQA to date. It has also been included in the multimodal multilingual evaluation benchmark IGLUE \cite{bugliarello2022IGLUE}. xGQA is the multilingual extension of the English GQA dataset \cite{hudson2018gqa} to 7 typologically diverse languages.

In this work, we use and empirically compare two state-of-the-art pretrained multimodal multilingual Transformer architectures: \textbf{M3P}~\cite{huang2020m3p} and \textbf{UC2}~\cite{zhou2021uc}.\footnote{For technical details of the two models, we refer the reader to their respective papers.} The standard cross-lingual \textit{zero-shot} transfer setup for VQA involves fine-tuning all the weights of the pretrained model on the downstream task data in the source language only. In the \textit{few-shot} setup, after the source-language fine-tuning, the model is additionally optimized on a handful of task-annotated examples in the target language \cite{pfeiffer2021xgqa}.

\section{Modeling Methods}
\label{s:modeling}

\sparagraph{Motivation}
Recent work on VQA in cross-lingual settings~\cite{pfeiffer2021xgqa,bugliarello2022IGLUE} benchmarked standard multimodal architectures in zero-shot and few-shot transfer scenarios on the xGQA dataset, without aiming to provide a deeper understanding of the particulars of the cross-lingual VQA task. At the same time, they report large gaps of cross-lingual transfer performance when compared to monolingual English performance, suggesting that there is ample room for improvement. In this work, we aim to leverage novel insights into different aspects of the cross-lingual VQA task (e.g., analyses over different question types or classification architectures) to guide improved cross-lingual VQA methods. In particular, we assess the impact of three orthogonal directions: \textbf{1)} classification architectures (\S\ref{ss:prediction}); \textbf{2)} (richer) input signals (\S\ref{sec:qtype}); \textbf{3)} fine-tuning strategies (\S\ref{ss:ftuning}).

\subsection{Classification Architecture Variants}
\label{ss:prediction}
The original work on xGQA~\cite{pfeiffer2021xgqa} evaluated only a simple `shallow' linear classification head, termed \textbf{Linear} here: the output [CLS] token of the pretrained Transformer-based model (which has cross-attended over all text and image features) is simply passed into a linear classification head. However, we hypothesize that this choice might have a substantial impact on transfer performance. Therefore, in the so-called \textbf{Deep} variant, instead of a linear classification head, we add a 2-layer transformation network ($f_{\textrm{trans}}$) with the GELU activation function~\cite{hendrycks2020gaussian}, dropout and a layer-normalization layer, before feeding the representations into a linear layer for classification. The first layer of $f_{\textrm{trans}}$ uses an orthogonal initializer~\cite{Saxe2014ExactST}. Unless noted otherwise, all of our following experiments are based on this `deeper' architecture; we illustrate the architecture in Figure~\ref{fig:arch} in Appendix~\ref{app:d}.

\subsection{Incorporating Inductive Bias into the Input Signal} 
\label{sec:qtype}

A large number of output classes (see \S\ref{s:background}) potentially amplifies the difficulty of zero-shot and few-shot cross-lingual transfer due to the need of aligning contextual representations in multiple languages for multi-class classifications.  Standard VQA datasets such as GQA and xGQA contain questions of five different structural types (\textit{Verify, Logical, Query, Choose, Compare}).\footnote{See Appendix~\ref{app:gqa_qtype} for example questions for each of the five question types.} \citet{pfeiffer2021xgqa} have demonstrated a considerable performance variation over different question types, e.g., there is a large cross-lingual performance drop especially for \textit{Choose}-type questions.

To help alleviate this issue, we propose to feed the model with designated question-type tokens~\cite{qtype,qtype2} which appear in GQA and xGQA. The idea is to influence the label distribution for the VQA classification task by conditioning on a question-type token.

More concretely, we prepend a question-type token $\texttt{QType}$ in English to the text input. We use structural question types as the question-type tokens; the text input then takes the following format: `\texttt{[QType]} : \texttt{[Question]}'.  

As the xGQA data contains questions with binary answers (i.e. \textit{Yes/No} questions). We anticipate that for a large fixed number of output classes, these questions should benefit the most from using the question-type tokens. The models which rely on this question-type conditioning are denoted with the superscript $^Q$, e.g., M3P$^{Q}$, see also later \S\ref{sec:notation}. 

Recent work~\citep{pet, li-liang-2021-prefix, Liu2021GPTUT, autoprompt} suggests that there exist more sophisticated prefixes/prompts and prompt-tuning methods. As our focus is not on conducting a large-scale analysis over different prompt-based conditioning, we leave this topic for future work.

\begin{figure}[t!]
  \centering
   \includegraphics[scale=0.45]{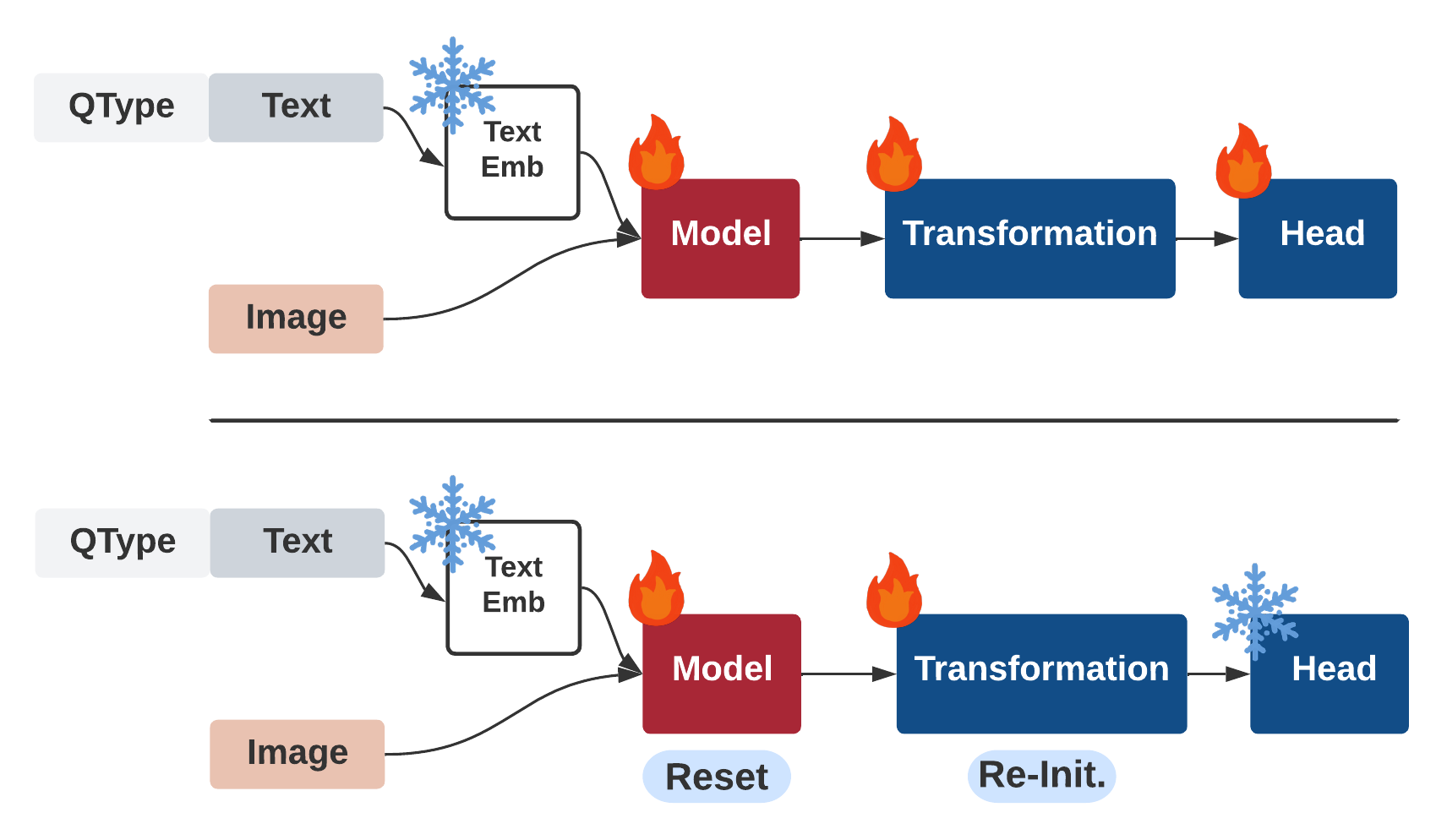}
   \vspace{-0.5mm}
    \caption{Self-Bootstrapping (\S\ref{ss:ftuning}). \textit{Top:} Fine-tuning with frozen text embeddings (Stage 1). \textit{Bottom:} Fine-tuning with text embeddings and classification head  frozen. Other parameters are reset to their pretrained values or randomly initialized (Stage 2).}
    \label{fig:sboot}
    \vspace{-1mm}
\end{figure}

\subsection{Fine-Tuning Strategy}
\label{ss:ftuning}
Misalignment of multilingual text embeddings \cite{Sogaard:2018acl,Dubossarsky:2020emnlp} has been indicated by \citet{pfeiffer2021xgqa} as one of the principal causes for reduced zero-shot performance in the cross-lingual VQA tasks. Therefore, we propose two fine-tuning strategies, tailored exactly towards mitigating such undesired shifts in the multilingual embedding space.

\rparagraph{Freezing Text Embeddings} 
In the first variant, we freeze text embeddings during fine-tuning and only optimize the Transformer weights and the classification head. This should prevent misalignment of the text embedding space during fine-tuning, as only the alignments between image and text embeddings change, but not text-to-text alignments. This strategy, labeled \textbf{+FT}, is referred to as contrastive tuning by~\citet{LitTune}.

\rparagraph{Self-Bootstrapping} 
Zero-shot cross-lingual transfer via standard fine-tuning is known to be sensitive to parameter initialization~\cite{bugliarello2022IGLUE}. Previous work has shown that fine-tuning a classification head first, then fine-tuning the model can effectively improve the generalization of the model~\cite{kumar2022finetuning, parameffhead}. Motivated by these insights from prior research, we first train the network to learn the classification head, then reset and fine-tune the remaining model parameters. This leads to a two-stage fine-tuning process, termed \textit{self-bootstrapping} (labeled \textbf{+SB}), illustrated in Figure~\ref{fig:sboot} and outlined here:

\vspace{0.8mm}
\noindent\emph{{Stage One:}} We fine-tune all parameters (with text embeddings frozen) on the task data.

\vspace{0.8mm}
\noindent\emph{{Stage Two:}} 
We \textbf{1)} freeze the classification head (excluding the bias parameters) and text embeddings, \textbf{2)} reset the remaining parameters in the multimodal multilingual model to pretrained weights, and \textbf{3)} re-initialize the $f_{\textrm{trans}}$ network (see \S\ref{ss:prediction}). We then fine-tune the transformer weights on the task data.\footnote{In our preliminary experiments, we found that self-bootstrapping-based fine-tuning still achieves better performance even if we perform Stage 1 with tunable text embeddings (i.e., standard fine-tuning). Freezing text embeddings in Stage 1 is an empirical decision, freezing them in Stage 2 is essential for self-bootstrapping to work.}

\vspace{0.5mm}
In order to make fair comparisons between +FT fine-tuning and self-bootstrapping, we define two extra +FT variants that match the fine-tuning budget of self-bootstrapping. In +FT$_{short}$ we fine-tune until the budget of self-bootstrapping's Stage 1 is matched. In +FT$_{long}$, we fine-tune until the total training budget of self-bootstrapping is matched. 

\subsection{Model Configurations and Notation}
\label{sec:notation}
Different choices across the orthogonal axes of classification architecture, input, and fine-tuning strategy give rise to a wide spectrum of \textit{model configurations}. In particular, we can independently choose \textbf{1)} between the Linear or Deep classification architecture; \textbf{2)} whether to include the information on the question type at input ($^Q$) or not; \textbf{3)} whether to apply standard fine-tuning from prior work \cite{pfeiffer2021xgqa}, or rely on +FT or +SB fine-tuning strategies. On top of this, we can also vary \textbf{4)} the underlying model (M3P or UC2), and \textbf{5)} the transfer scenario (zero-shot versus few-shot). For clarity of presentation, unless noted otherwise, we always assume zero-shot scenarios and Deep classification architecture. Moreover, different variants are also labeled in a systematic manner using abbreviations introduced in \S\ref{ss:prediction}-\S\ref{ss:ftuning}: e.g., \textit{M3P+SB} means that we apply self-bootstrapping on the underlying M3P model (with Deep architecture assumed). In another example, \textit{UC2$^{Q}$+FT$_{long}$} means that we apply the \textit{long} variant of +FT fine-tuning (see \S\ref{ss:ftuning}) with UC2 as the underlying model, and we condition the model on the information about question types.

\section{Analysis Methods}
\label{sec:analysis_methods}

The VQA task is inherently multimodal---a model is required to reason over both images and questions in textual form to solve the task. However, as with some unimodal text-only tasks \cite{gururangan-etal-2018-annotation,Poliak:2018sem} VQA models might also be prone to `taking shortcuts', that is, exploiting spurious correlations and artifacts of data creation. In other words, the VQA model could circumvent the multimodal aspect and only focus on a single modality to solve the task~\cite{agrawal-etal-2016-analyzing, Agrawal2018DontJA}. Therefore, to better understand the multimodal reasoning abilities of VQA models in cross-lingual transfer, we propose several diagnostic approaches and methods that ablate the input features of the models, inspired by the diagnostic methods of ~\citet{frank-etal-2021-vision} and ~\citet{shrestha-etal-2020-negative} in monolingual setups. They should provide us with deeper insights into the inner workings of cross-lingual VQA models.

\subsection{Unimodal \textit{Evaluation}}
The first set of analyses involves a combination of standard multimodal (\textbf{MM}) training with unimodal inference/evaluation. During training, we pass both visual features and text tokens into the model. However, at inference, we provide the model with features of only one modality (Visual modality: \textbf{V} or Text: \textbf{T}). This naturally gives rise to the following two experimental setups:

\vspace{0.6mm}
\noindent{\textbf{MM-V}}: When evaluating on xGQA's test set, we pass only a single `?' as textual input to the model, while the standard visual features are used.

\vspace{0.6mm}
\noindent\textbf{MM-T}: At inference, we zero out all visual features (e.g., object features, spatial features), only providing the model with the total number of objects detected; the unchanged questions in the textual form are provided to the model.

\subsection{Unimodal \textit{Training} and \textit{Evaluation}}
\label{ss:unimodal-te}
Next, we probe purely unimodal models \textit{trained} on a single modality (\textbf{V} or \textbf{T}):  during training, the model is provided only with visual features or text tokens; at inference, we again only provide the model with unimodal features from the same modality. This creates three experimental setups:

\vspace{0.6mm}
\noindent{\textbf{V-V}}: We pass only `?' as a (placeholder) textual input to the model, while the standard visual features (from the full multimodal model) are used.

\vspace{0.6mm}
\noindent\textbf{T-T}: All visual features are zeroed out; we only provide the number of objects detected; the unchanged questions in the textual form are provided. 

\vspace{0.6mm}
\noindent\textbf{T$^G$-T$^G$}: We randomly sample object features from a Gaussian distribution with a mean and a standard deviation that match the actual object feature distribution for that image. Spatial features and the number of objects detected are kept as in the full MM model. The standard unchanged questions in the textual form are provided to the model. 

\begin{table*}[!t]
\setlength\tabcolsep{2pt}
      \resizebox{0.99\textwidth}{!}{
    \centering
    \scriptsize
    \begin{tabular}{c l c ccccccc c}
    \toprule
        & \textbf{Method} & En & De & Zh & Ko & Id & Bn & Pt & Ru & \textbf{Avg}\\
    \cmidrule(lr){2-2} \cmidrule(lr){3-3} \cmidrule(lr){4-11} 
    G1    &\mmmp/$^{*}$ (Linear) & 
        51.88$\scriptscriptstyle\pm0.7$ &
        27.45$\scriptscriptstyle\pm5.8$ &
        16.33$\scriptscriptstyle\pm8.3$ &
        13.70$\scriptscriptstyle\pm5.4$ &
        25.25$\scriptscriptstyle\pm11.4$ &
        10.59$\scriptscriptstyle\pm3.4$ &
        21.10$\scriptscriptstyle\pm3.4$ &
        20.95$\scriptscriptstyle\pm3.3$ &
        19.34
        \\
        &\mmmp/$^{*}$ & 
        51.66$\scriptscriptstyle\pm0.6$ &
        35.33$\scriptscriptstyle\pm5.4$ &
        27.80$\scriptscriptstyle\pm10.9$ &
        25.55$\scriptscriptstyle\pm11.4$ &
        30.54$\scriptscriptstyle\pm9.8$ &
        17.94$\scriptscriptstyle\pm8.6$ &
        30.61$\scriptscriptstyle\pm7.2$ &
        29.74$\scriptscriptstyle\pm6.6$ &
        28.22 \\
         \cmidrule(lr){2-2} \cmidrule(lr){3-3} \cmidrule(lr){4-11} 
            G2 & \mdoq/ & 
        50.90$\scriptscriptstyle\pm0.5$ &
        37.95$\scriptscriptstyle\pm1.5$ &
        35.06$\scriptscriptstyle\pm2.6$ &
        32.31$\scriptscriptstyle\pm3.4$ &
        36.56$\scriptscriptstyle\pm2.0$ &
        27.69$\scriptscriptstyle\pm1.8$ &
        36.64$\scriptscriptstyle\pm2.4$ &
        37.30$\scriptscriptstyle\pm4.6$ &
        34.79 \\
         \cmidrule(lr){2-2} \cmidrule(lr){3-3} \cmidrule(lr){4-11} 
        G3 & M3P + SB &
        47.26$\scriptscriptstyle\pm1.0$ &
        35.71$\scriptscriptstyle\pm6.1$ &
        29.70$\scriptscriptstyle\pm8.2$ &
        30.33$\scriptscriptstyle\pm8.3$ &
        28.16$\scriptscriptstyle\pm2.7$ &
        20.70$\scriptscriptstyle\pm3.9$ &
        34.65$\scriptscriptstyle\pm6.5$ &
        34.63$\scriptscriptstyle\pm6.9$ &
        30.56\\
         \cmidrule(lr){2-2} \cmidrule(lr){3-3} \cmidrule(lr){4-11} 
        G4 & M3P$^{Q}$ + FT$_{short}$ & 
        49.48$\scriptscriptstyle\pm0.3$ &
        38.68$\scriptscriptstyle\pm2.6$ &
        34.94$\scriptscriptstyle\pm2.2$ &
        34.17$\scriptscriptstyle\pm2.6$ &
        \textbf{37.18}$\scriptscriptstyle\pm2.4$ &
        \textbf{30.00}$\scriptscriptstyle\pm2.2$ &
        37.35$\scriptscriptstyle\pm1.9$ &
        37.57$\scriptscriptstyle\pm2.4$ &
        35.56\\
        & M3P$^{Q}$ + FT$_{long}$ & 
        51.00$\scriptscriptstyle\pm0.9$ &
        38.42$\scriptscriptstyle\pm2.1$ &
        35.05$\scriptscriptstyle\pm2.1$ &
        33.38$\scriptscriptstyle\pm2.5$ &
        36.24$\scriptscriptstyle\pm2.3$ &
        27.77$\scriptscriptstyle\pm1.7$ &
        36.78$\scriptscriptstyle\pm2.3$ &
        37.42$\scriptscriptstyle\pm2.0$ &
        35.01\\
       & \gbc\msboot / & 
        \gbc46.70$\scriptscriptstyle\pm0.7$ 
        & \gbc\textbf{39.52}$\scriptscriptstyle\pm1.3$ 
        & \gbc\textbf{36.15}$\scriptscriptstyle\pm0.9$ 
        & \gbc\textbf{35.67}$\scriptscriptstyle\pm1.1$
        & \gbc36.73$\scriptscriptstyle\pm1.6$
        & \gbc29.75$\scriptscriptstyle\pm1.4$
        & \gbc\textbf{37.59}$\scriptscriptstyle\pm0.8$
        & \gbc\textbf{37.93}$\scriptscriptstyle\pm0.9$
        & \gbc\textbf{36.19}\\
        
        \midrule
        
   G1     &\ucc/$^{*}$ (Linear) & 
        57.83$\scriptscriptstyle\pm	0.3$ &
        40.57$\scriptscriptstyle\pm	1.7$ &
        35.54$\scriptscriptstyle\pm	3.4$ &
        16.95$\scriptscriptstyle\pm	6.1$ &
        34.18$\scriptscriptstyle\pm	0.8$ &
        8.53$\scriptscriptstyle\pm	1.9$ &
        24.90$\scriptscriptstyle\pm	3.7$ &
        24.05$\scriptscriptstyle\pm	4.6$ &
        26.39	 
        \\
        &\ucc/$^{*}$  &
        58.31$\scriptscriptstyle\pm0.2$ &
        41.33$\scriptscriptstyle\pm1.6$ &
        34.77$\scriptscriptstyle\pm2.2$ &
        23.87$\scriptscriptstyle\pm1.5$ &
        34.79$\scriptscriptstyle\pm1.3$ &
        11.82$\scriptscriptstyle\pm1.9$ &
        29.30$\scriptscriptstyle\pm4.5$ &
        29.41$\scriptscriptstyle\pm3.7$ &
        29.33
       \\
  \cmidrule(lr){2-2} \cmidrule(lr){3-3} \cmidrule(lr){4-11} 

   G2  & \ucdoq/ & 
        58.35$\scriptscriptstyle\pm0.4$ &
        45.13$\scriptscriptstyle\pm0.8$ &
        42.85$\scriptscriptstyle\pm0.9$ &
        31.33$\scriptscriptstyle\pm1.0$ &
        35.64$\scriptscriptstyle\pm0.9$ &
        24.86$\scriptscriptstyle\pm0.6$ &
        37.19$\scriptscriptstyle\pm0.6$ &
        38.61$\scriptscriptstyle\pm0.9$ &
        36.52 \\
\cmidrule(lr){2-2} \cmidrule(lr){3-3} \cmidrule(lr){4-11} 
        G3& UC2 + SB &    
        58.52$\scriptscriptstyle\pm0.4$ &
        48.51$\scriptscriptstyle\pm1.3$ &
        43.97$\scriptscriptstyle\pm0.3$ &
        35.08$\scriptscriptstyle\pm2.0$ &
        37.33$\scriptscriptstyle\pm3.2$ &
        19.09$\scriptscriptstyle\pm4.5$ &
        35.29$\scriptscriptstyle\pm2.9$ &
        35.99$\scriptscriptstyle\pm3.5$ &
        36.46\\
        \cmidrule(lr){2-2} \cmidrule(lr){3-3} \cmidrule(lr){4-11} 
       G4 & UC2$^{Q}$ + FT$_{short}$ & 
        57.83$\scriptscriptstyle\pm0.5$ &
        47.17$\scriptscriptstyle\pm1.6$ &
        45.59$\scriptscriptstyle\pm0.9$ &
        34.19$\scriptscriptstyle\pm0.7$ &
        37.04$\scriptscriptstyle\pm1.1$ &
        24.94$\scriptscriptstyle\pm0.5$ &
        38.32$\scriptscriptstyle\pm1.2$ &
        39.96$\scriptscriptstyle\pm1.4$ &
        38.17 \\
        & UC2$^{Q}$ + FT$_{long}$  & 
        58.15$\scriptscriptstyle\pm	0.6$ &
        44.27$\scriptscriptstyle\pm0.5$ &
        42.49$\scriptscriptstyle\pm0.4$ &
        29.75$\scriptscriptstyle\pm0.3$ &
        36.81$\scriptscriptstyle\pm0.4$ &
        24.48$\scriptscriptstyle\pm0.2$ &
        35.39$\scriptscriptstyle\pm	0.4$ &
        37.32$\scriptscriptstyle\pm0.4$ &
        35.79 \\
        & \gbc\ucsboot/ &  
        \gbc58.57$\scriptscriptstyle\pm0.2$ &
        \gbc\textbf{49.51}$\scriptscriptstyle\pm1.1$ &
        \gbc\textbf{46.52}$\scriptscriptstyle\pm0.9$ &
        \gbc\textbf{36.48}$\scriptscriptstyle\pm1.3$ &
        \gbc\textbf{38.92}$\scriptscriptstyle\pm1.3$ &
        \gbc\textbf{26.23}$\scriptscriptstyle\pm	1.5$ &
        \gbc\textbf{39.76}$\scriptscriptstyle\pm0.6$ &
        \gbc\textbf{41.72}$\scriptscriptstyle\pm0.3$ &
        \gbc\textbf{39.87} \\
    \bottomrule
    \end{tabular}
    }
    \caption{Zero-shot transfer results on xGQA. Avg. refers to the average accuracy across languages excluding English. Group G1: baselines. $*$: our runs of baselines trained on balanced GQA. Group G2: results using a question-type token. Group G3: results using self-bootstrapping (+SB). Group G4: combining different fine-tuning strategy with the use of question-type tokens. Best results in each column and per each pretrained model across Groups G1-G4 are shown in \textbf{bold}. Results are averaged across four random seeds.}
    \label{tab:main_xgqa}
\end{table*}

\section{Experimental Setup}
\label{sec:exp}
\sparagraph{Pretrained Models and Data}
As introduced in \S\ref{s:background}, we 1) rely on two standard state-of-the-art multimodal multilingual transformers~\cite[\mmmp/, \ucc/;][]{huang2020m3p,zhou2021uc} as the underlying pretrained models, and 2) conduct all evaluations on the standard monolingual English GQA dataset, and its multilingual extension: xGQA.

The GQA dataset consists of two training sets: \textbf{full} and \textbf{balanced}. The full dataset contains 113K images and 22M questions, whereas the balanced dataset consists of 1.7M data samples. The dataset also contains a balanced test-dev set with 12,578 questions and 398 images for evaluation. In xGQA, the questions are manually translated from the GQA test-dev set into 7 different languages: Bengali, Chinese (simplified), German, Indonesian, Korean, Portuguese, and Russian. xGQA provides a zero-shot evaluation set and a different training/evaluation set for the few-shot setting. Please see the original paper for details.

\rparagraph{Training Details and Hyperparameters}
Following the recommendations from \citet{bugliarello2022IGLUE}, we predominantly run training on the more lightweight \emph{balanced} subset of GQA.\footnote{Another established yet less efficient training procedure is to train on the full GQA dataset first, then further train on the balanced dataset~\cite{Li2020OscarOA}. This procedure can produce good results on the English evaluation dataset at the cost of a substantial increase in computation demands (${\sim}4$ days on one NVIDIA V100 for one model). Furthermore, our initial experiments have indicated that training with the balanced set performs similarly to the previously reported baselines in the xGQA paper while using substantially less computing. We stress that we also further run experiments under the more demanding training regime~\cite{Li2020OscarOA} with the best-performing model configuration from our experiments. For more details, we refer the reader to \S\ref{sec:further_exp}.} We also define a total training budget of 6 epochs (less than 24 hours of training). For the self-bootstrapping procedure, this means the total training time (Stage 1 + Stage 2) is equal to 6 epochs. See Appendix~\ref{app:hpm} for further details.

\section{Results and Discussion}

In \S\ref{sec:impact_modelling_methods}, we discuss the results of the different modeling approaches across the three dimensions (see \S\ref{s:modeling}): classification architectures, input signals and fine-tuning strategies. A finer-grained analysis concerning different structural question types is provided in \S\ref{sec:structural_q_type_analysis}. Finally, in \S\ref{sec:analysis_ablation} we delve deeper into the VQA models' susceptibility towards exploiting unimodal biases and artifacts of the VQA datasets, relying on model variants discussed in \S\ref{sec:analysis_methods}.

\subsection{Impact of Modeling Methods}
\label{sec:impact_modelling_methods}

A summary of the results with a wide spectrum of possible model configurations (see \S\ref{sec:notation}) is provided in Table~\ref{tab:main_xgqa}, with accuracy as the main metric.

First, an interesting trend emerges: different model configurations have \textit{no} significant effect on performance in the source language (English), especially so for the better-performing pretrained model UC2. However, variations in different modeling choices from \S\ref{s:modeling} do show \textit{considerable} impact on cross-lingual transfer performance: we report gains by more than 16 and 13 absolute accuracy points for M3P and UC2, respectively.

\vspace{1.6mm}
\noindent\textbf{Classification Architectures.} Surprisingly, simply adding additional non-linear layers to the prediction head has a considerable impact on the cross-lingual transfer performance of the baseline models (especially for the M3P model) while performance in the source language stays nearly the same (Table~\ref{tab:main_xgqa}, Group G1). Put simply, a deeper classification architecture seems to benefit cross-lingual transfer performance, and the extent of its impact cannot be captured by monolingual English-only evaluation. Further, to isolate the source of these improvements, we conducted additional experiments by removing the layer normalization component from the deeper architecture. The results are provided in Table~\ref{app:tab_ln} in the Appendix. 
Another key observation is that the impact of depth is model-dependent with stronger configurations. While it yields large gains when we start from the baseline transfer models (G1), the gains from the classification architecture are less pronounced or even non-existent, e.g., for the best-performing \ucsboot/ model variant (see Group G4): 39.87 (Deep) versus 40.89 (Linear). The gains from classification architecture remained for the \mmmp/ model variant: 36.19 (Deep) versus 18.24 (Linear).

\vspace{1.6mm}
\noindent\textbf{Input Signal.} The large number of output classes of GQA potentially results in a noisy distribution over the predicted labels when sentences in a different language are passed into the model. We find that including the question-type token (Q) improves the average cross-lingual zero-shot transfer accuracy by more than 10\% relatively for both \mdoq/ and \ucdoq/ (Table~\ref{tab:main_xgqa}, Group G2). This modeling decision again has an inconsequential impact on the source language but suggests that the question-type token can partially mitigate the poor performance of cross-lingual transfer. A comparison of G3 versus G4 models in Table~\ref{tab:main_xgqa} demonstrates that including the question type at input yields gains of almost 6 accuracy points with M3P, and more than 3 points with UC2, with especially large gains for Bengali as the lowest-performing language.

\vspace{1.6mm}
\noindent\textbf{Fine-tuning Strategy.}
Freezing the embeddings to mitigate a shift in the multilingual embedding space results in positive gains for cross-lingual scenarios (Table~\ref{tab:main_xgqa}, Group G4). The self-bootstrapping strategy (+SB with and without Q) achieves further gains over both +FT embedding-freezing experimental setups. At the same time, it also yields much lower variance across languages (with Q). This validates that resetting parameters with self-bootstrapping positively impacts model performance, and supports our hypothesis that first fixing the classifier weights to good values leads to better performance and lower variance. Note that the average +SB results of \ucc/ are statistically significant against \ucdoq/ and \ucdoq/ +FT$_{short}$ ($p < 0.05$).

\begin{figure*}[!htbp]
    \centering
    \begin{subfigure}[t]{0.49\textwidth}
        \includegraphics[clip, trim=0.7cm 0.65cm 0.2cm 0.65cm, width=\textwidth]{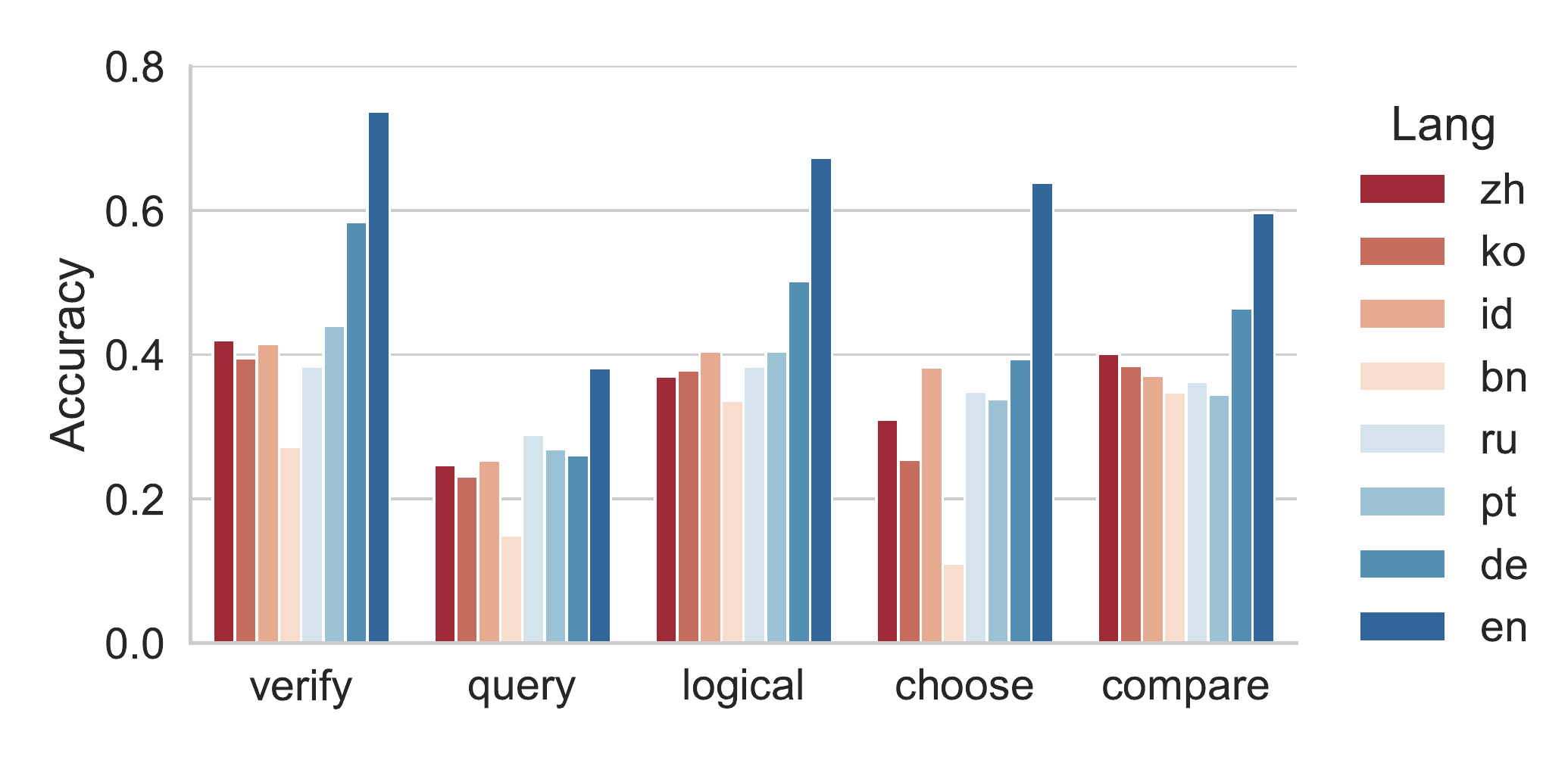}
        \caption{M3P  }%
        \label{fig:analysis_m3p}
    \end{subfigure}
    \hfill
    \begin{subfigure}[t]{0.49\textwidth}
        \includegraphics[clip, trim=0.7cm 0.65cm 0.2cm 0.65cm,width=\textwidth]{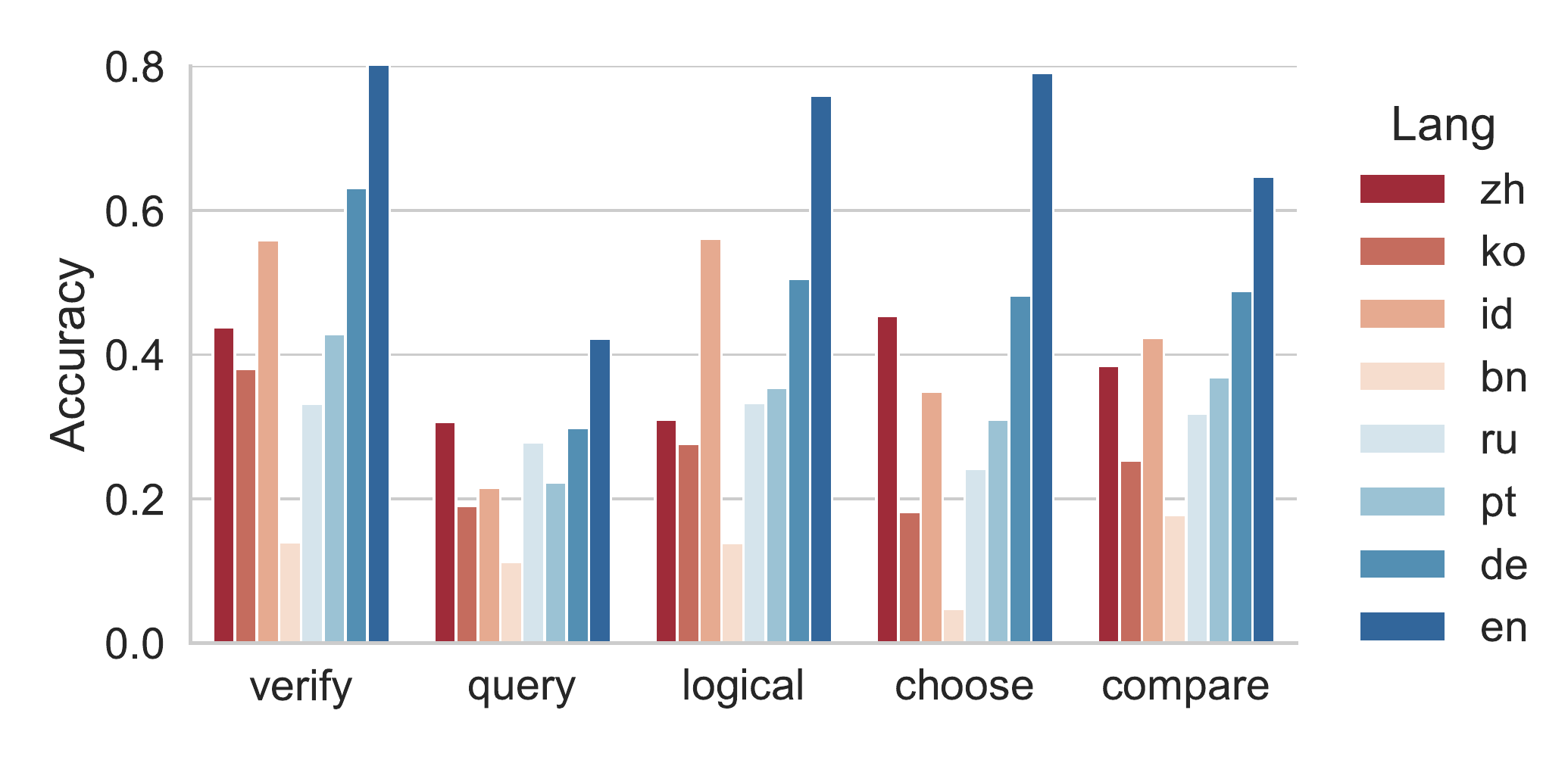}
        \caption{UC2 }%
        \label{fig:analysis_uc2}
        
    \end{subfigure}
    
    \begin{subfigure}[t]{0.49\textwidth}
        \includegraphics[clip, trim=0.7cm 0.65cm 0.2cm 0.65cm,width=\textwidth]{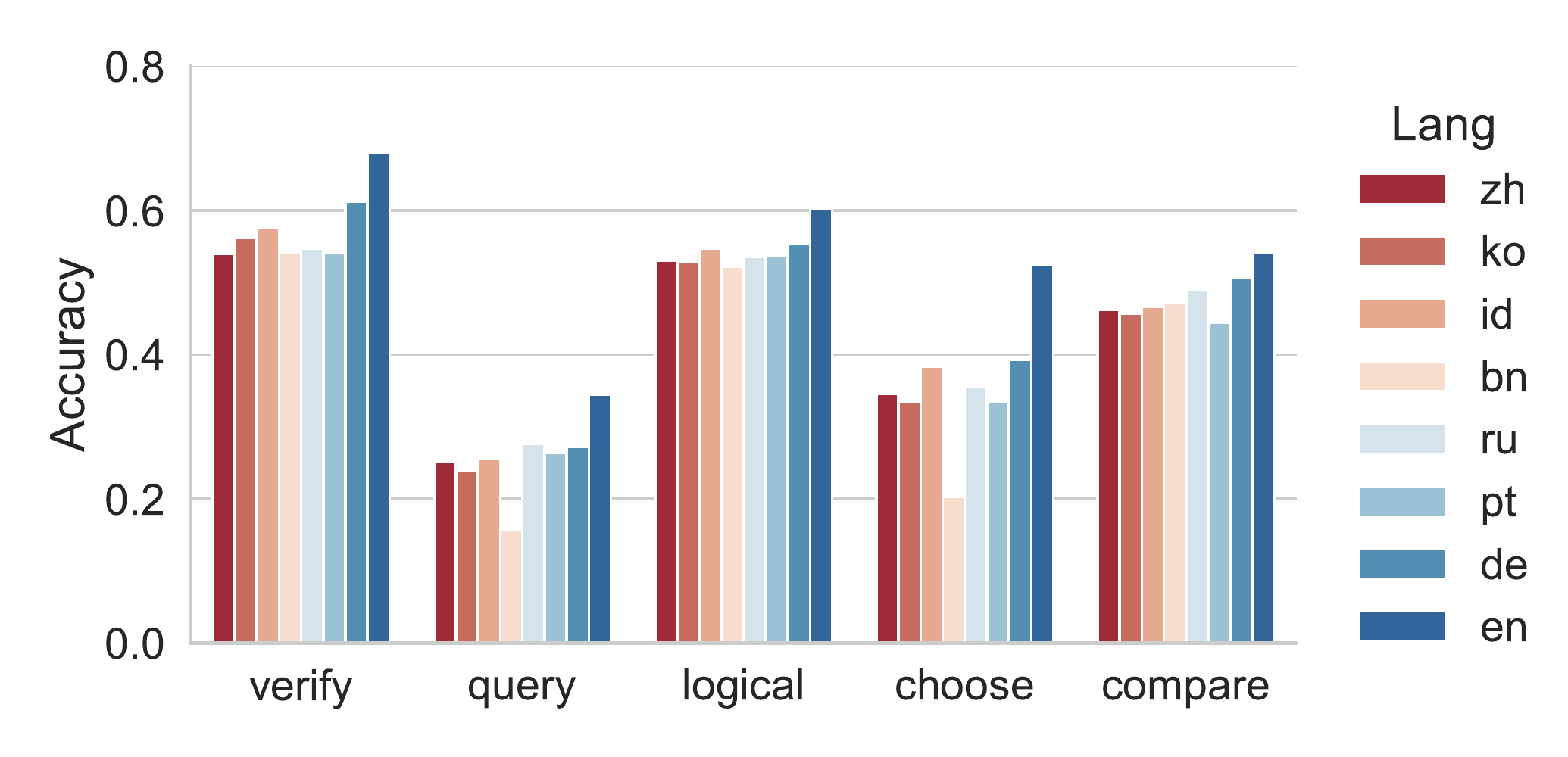}
        \caption{\msboot/}
        \label{fig:analysis_sboot_m3p}
    \end{subfigure}
    \begin{subfigure}[t]{0.49\textwidth}
        \includegraphics[clip, trim=0.7cm 0.65cm 0.2cm 0.65cm,width=\textwidth]{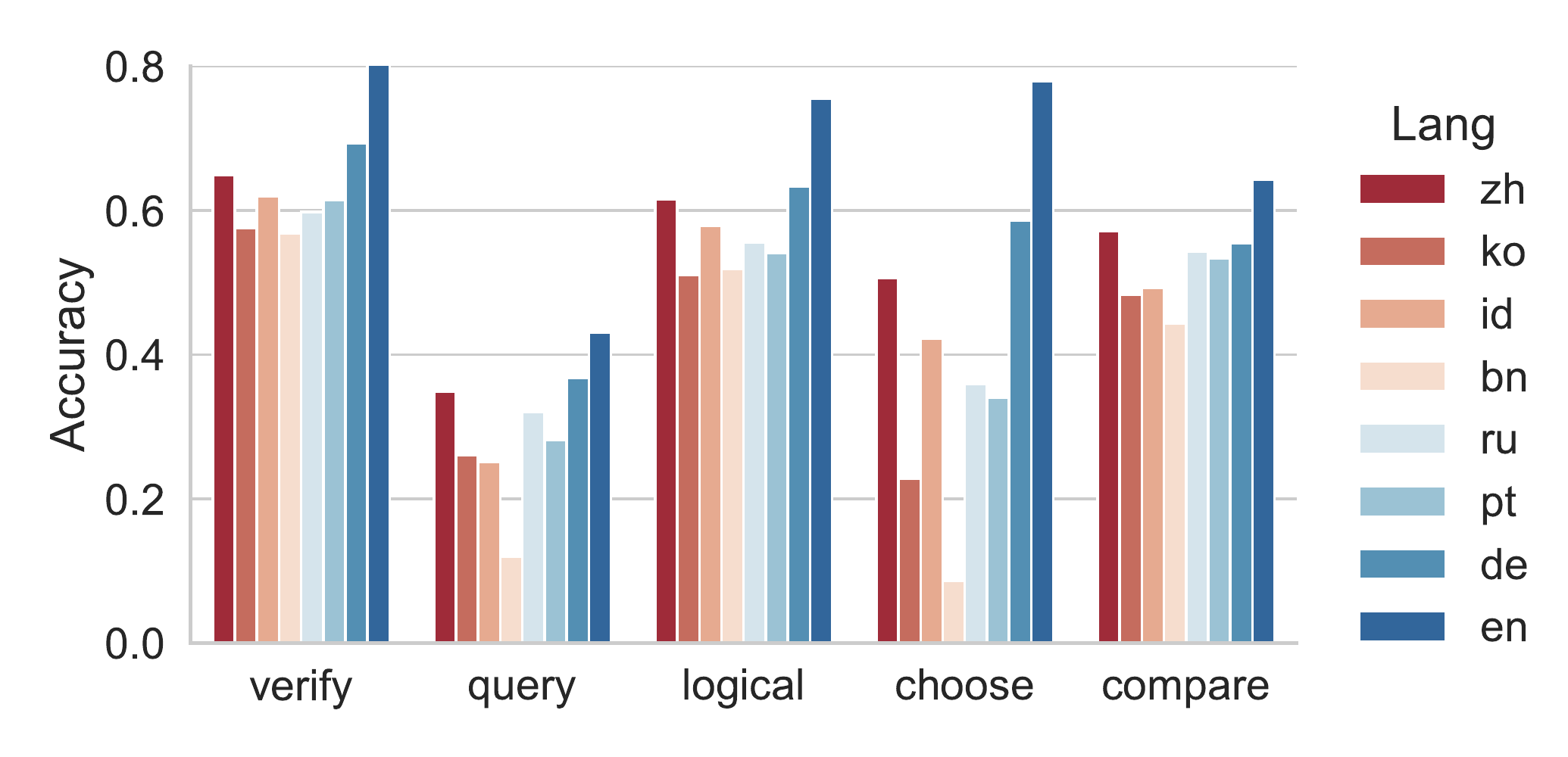}
        \caption{\ucsboot/}
        \label{fig:analysis_sboot_uc2}
    \end{subfigure}
    \caption{Zero-shot cross-lingual transfer performance across individual question types in GQA and xGQA. }
    \label{fig:structural_q_types}
\end{figure*}

\vspace{1.6mm}
\noindent\textbf{Performance across Question Types.}
\label{sec:structural_q_type_analysis} Finer-grained results per individual question type are summarized in Figure~\ref{fig:structural_q_types}, where we compare the baseline models with the best-performing variant, which utilizes the question-type at the input and the self-bootstrapping strategy. In sum, we observe gains across all structural question-types for such $^Q$+SB model configurations, both for M3P and UC2. Performance on \textit{Query} and \textit{Choose} questions meets substantial gains, suggesting that improving the alignment between multilingual text embeddings has a positive effect on performance, especially for non-binary, free-form question-types. Complete results are available in Appendix~\ref{app:gqa_qtype}).

\subsection{Learning Biases: Multi-Modal versus Unimodal VQA?}\label{sec:analysis_ablation}

We use the analysis methods from \S\ref{sec:analysis_methods} to determine whether the
underlying models have learned to rely on a single modality to make predictions, either due to spurious correlations in the data or the model's inability to effectively combine multi-modal features. The main results are provided in Table~\ref{tab:modality}.

\rparagraph{Unimodal \textit{Evaluation}} The scores of MM-T/MM-V ablations reveal the sensitivity to missing features in each input modality at test time. We observe a drop in accuracy of more than 50\%  across all question types in the MM-T/MM-V experiments compared to their counterparts that assume `full-feature' multi-modal input at inference. Moreover, \textit{Verify}, \textit{Logical} and \textit{Compare} questions seem more dependent on text features. The results confirm that the trained model needs both modalities to achieve good cross-lingual performance, although not at equal proportions. In other words, high zero-shot transfer performance observed in our experiments are obtained by leveraging both modalities in synergy, and not by `taking unimodal shortcuts' (\S\ref{sec:analysis_methods}).

\rparagraph{Unimodal \textit{Training} and \textit{Evaluation}} V-V/T-T/T$^G$-T$^G$ experiments reveal the worst-case exploitation of the data biases in modalities by the models. The results suggest that a majority of the final performance can be attained with text features in fine-tuned models for the \textit{Logical}, \textit{Verify}, and \textit{Compare} question types. Therefore, the results indicate that these question types contain modality biases that can be exploited by unimodal VQA architectures. The exploitable data biases could also explain the observations from prior experiments. We suspect this could also explain the asymmetrical attention over modalities, observed by \citet{frank-etal-2021-vision} in monolingual multi-modal models.

\rparagraph{Biases across Question Types.}
 Unimodally trained models can only attain $\sim$20\%  (M3P) and $\sim$26\% (UC2) accuracy at best for the \textit{Query} question type, with similar trends observed for \textit{Choose}. Exposing the models to increasingly more visual features (from T-T over T$^G$-T$^G$ to the full multi-model) yields significant performance gains. It thus indicates that \textit{Query} and \textit{Choose} questions contain fewer exploitable data biases, and additional image-text grounding could help improve predictions. Previous work in monolingual settings~\cite{Kervadec2021RosesAR} concludes that \textit{Compare} and \textit{Query} questions should be focused on for future improvements. Here, in the cross-lingual setting, we found \textit{Query} and \textit{Choose} questions as the most difficult questions with the largest gaps in monolingual English performance.
 
 Table~\ref{tab:modality} also reveals that more sophisticated fine-tuning strategies such as self-bootstrapping, which prevent multilingual text embedding shifts, are an effective way to improve performance on these two (most challenging) question types.

\begin{table}[h!]
\def\arraystretch{0.999}
\setlength\tabcolsep{2pt}
\scriptsize
    \centering   
     \resizebox{0.48\textwidth}{!}{
    \begin{tabular}{l ccc cc cc}
    \toprule
    {\bf \mmmp/}  & V-V & T-T & T$^G$-T$^G$ & \mdoq/ & M3P$^Q$+SB & MM-V & MM-T\\
    \cmidrule(lr){1-1} \cmidrule(lr){2-4} \cmidrule(lr){5-6} \cmidrule(lr){7-8}
    Verify	& 45.19	&  53.98	&	54.88	& 58.35 & 55.98 & 0.1 & 18.59\\
    Logical	& 43.18  & 51.66 & 53.06 &  53.89 & 53.65 & 0.0 & 19.87 \\
    Compare	& 27.76	& 46.22 & 39.64 & 45.82 & 47.14 & 0.1 & 17.85 \\
    \gbc{Query} &  \gbc2.63	& \gbc4.39 & \gbc11.42 & \gbc21.86 & \gbc\textbf{24.50} & \gbc6.81 & \gbc4.46\\
    \gbc{Choose}  &  \gbc1.21	& \gbc8.52 & \gbc22.26 & \gbc29.43 & \gbc\textbf{33.57} & \gbc2.08 & \gbc12.22 \\
    \midrule
    {\bf \ucc/} & V-V & T-T &  T$^G$-T$^G$ &  \ucdoq/ & UC2$^Q$+SB & MM-V & MM-T\\
    \cmidrule(lr){1-1} \cmidrule(lr){2-4} \cmidrule(lr){5-6} \cmidrule(lr){7-8}
    Verify & 44.60 & 51.87 &	57.00 & 59.94 & 61.70 &   4.21 & 24.91\\
    Logical& 44.26  & 50.78 &	 52.57 & 54.87 & 56.49 &  6.27 & 21.12\\
    Compare	& 33.45 & 40.55 &	46.91 & 49.15 & 51.73 & 2.85 & 21.08  \\
    \gbc{Query} & \gbc3.39 & \gbc6.23 &	\gbc12.11  & \gbc23.94 &  \gbc\textbf{27.88} & \gbc7.30 & \gbc0.02 \\
    \gbc{Choose} & \gbc1.39 & \gbc17.24 &	\gbc23.76 & \gbc29.66 &  \gbc\textbf{36.14} & \gbc2.27 & \gbc0.14\\
    \bottomrule
    \end{tabular}
    } 
    \caption{Zero-shot transfer results of M3P$^{Q}$/UC2$^{Q}$ trained and tested with visual features only (V-V), text features only (T-T), text features with partial visual features (T$^G$-T$^G$), as well as of M3P$^{Q}$+SB/UC2$^{Q}$+SB trained using all features, but exposed only to visual features (MM-V) or text features (MM-T) at inference (\S\ref{sec:analysis_methods}). The scores are averaged over all target languages in xGQA, excluding English.}
    \label{tab:modality}
\end{table}

In summary, it is crucial to conduct such finer-grained analyses across different question types in the multilingual VQA tasks, and not treat them equally with only a global accuracy metric. In particular, our results render \textit{Query} and \textit{Choose} question types as by far the most challenging question types for cross-lingual transfer and the types that do not suffer from exploitable data biases. Future research in multilingual VQA should put more emphasis on such questions, and approaches that prevent the exploitation of unimodal data biases. Future research should also look beyond the question types currently covered by xGQA, and introduce even more challenging types.

\section{Additional Results}
\label{sec:further_exp}

\sparagraph{Training with Full English GQA} To validate the effectiveness of our approach in setups where more data in the source language is available, we additionally run experiments in another VQA setup: we train the best-performing method UC2$^Q$+SB for 5 epochs on the unbalanced English GQA dataset, followed by 2 epochs on the balanced dataset. Despite the fact that this variant leverages more source-language training data and consumes considerably more compute, we do not observe any gain on monolingual English performance, and observe only a small gain in the cross-lingual zero-shot setup: 
the accuracy score, averaged across all the target languages, increases from 39.87 to 40.51.\footnote{Table~\ref{tab:main_full} in Appendix~\ref{app:full} provides per-language accuracy.}

\begin{table}[t!]

    \scriptsize
    \centering

    \def\arraystretch{0.9}
\setlength\tabcolsep{2pt}

\begin{tabularx}{\columnwidth}{l YYYY}
\toprule
    \textbf{Method} & \multicolumn{1}{c}{0} & \multicolumn{1}{c}{1} & \multicolumn{1}{c}{5} & \multicolumn{1}{c}{48}  \\
 \cmidrule(lr){1-1}  \cmidrule(lr){2-2} \cmidrule(lr){3-5}
    \mmmp/ &  35.58 & 37.62 & 39.29 &	42.28 \\
    \mmmp/ + SB & 33.73& 35.89&	39.27&	42.46\\
    \mdoq/&  33.81 & 35.40 & 37.80	& 41.87\\
    \gbc\msboot/  & \gbc37.14 &	\gbc37.50 & \gbc38.16 & \gbc40.00 \\
 \midrule
    \ucc/ &  30.15 &  36.09 & 38.67 & 	44.37 \\
    \ucc/ + SB & 38.09  & 40.51 & 42.14 & 46.68\\
    \ucdoq/ & 37.28 & 39.24 & 40.88 & 45.11 \\
    \gbc\ucsboot/ & \gbc39.83 &	\gbc42.35& 	\gbc43.68 &	\gbc46.62\\
\bottomrule
    \end{tabularx}

    \caption{Averaged few-shot (0/1/5/48-shot) accuracy scores on xGQA (excl. English) for selected models.}
    \label{tab:fs}
\end{table}

\vspace{1em}
\rparagraph{Few-shot Experiments} 
Besides the zero-shot transfer scenario---which is the primary focus of this work---we also evaluate whether similar findings extend to few-shot scenarios, where a handful of annotated examples in the target language is assumed. Following the standard setup of~\citet{lauscher-etal-2020-zero} we start from the weights of the best-performing model, already fine-tuned on English VQA data. We then further fine-tune it on the few examples in the target language. In particular, we conduct few-shot experiments with 1, 5, and 48 images.\footnote{We choose 1 and 5 shots because these are typical in few-shot training setups \cite{zhao-etal-2021-closer}. 48 shots are the maximum available training data for the few-shot evaluation.}  Following \citet{pfeiffer2021xgqa} we fine-tune for 10 epochs, with a learning rate of 5e-5.

The results are summarized in Table~\ref{tab:fs} (see Table~\ref{tab:app_few_shot} in Appendix~\ref{app:few_shot} for full results), and indicate two key findings. First, we corroborate findings from prior work, where it was shown that fine-tuning on an increasing number of shots/examples in the target language generally improves model performance. Second, although baseline models are able to recover more performance from zero-shot to few-shot setups, our best-performing configuration with UC2 still significantly outperforms the baseline. We attribute the on-par performance across M3P variants to M3P's sensitivity to initialization and high variance. These results indicate that few-shot fine-tuning is an \textit{additional} cost-efficient approach, orthogonal to our modeling enhancements from \S\ref{s:modeling}, to further improve VQA model performance in the target language.

\section{Related Work}

Transformer-based models trained on multimodal data~\cite[][\textit{ inter alia}]{tan-bansal-2019-lxmert, Li2020OscarOA,vlt5,Shen2021CLIPVIL, Kamath_2021_ICCV} have demonstrated impressive results on English-only VQA tasks. 
However, as training and evaluation data has previously only been available in high resource languages \cite{elliott-etal-2016-multi30k,elliott-etal-2017-findings,barrault-etal-2018-findings, are-you-talking-to-machines}, progress in multilingual vision-and-language learning has not kept pace. 

More comprehensive multilingual multimodal benchmarks  have been developed only recently \cite[\textit{inter alia}]{Srinivasan2021WIT,su-etal-2021-gem,liu-etal-2021-visually,pfeiffer2021xgqa,Wang2021MultiSubs, bugliarello2022IGLUE} making it possible to evaluate multimodal models which have either been pretrained on multilingual data  \cite{huang2020m3p,zhou2021uc} or extended to unseen languages \cite{liu-etal-2021-visually,pfeiffer2021xgqa}. 

Our work complements this recent line of work by delving deeper into cross-lingual visual question answering, again highlighting the inherent difficulty of multilingual multimodal learning.

\section{Conclusion}
 In this work, we provide an extensive analysis of the issues present in VQA-related multilingual vision-and-language learning, aiming to inspire new solutions that can improve cross-lingual VQA performance. To this end, we studied simple yet effective methods that increase previously low transfer performance and thus substantially reduce the gap to monolingual English performance. This has been achieved through more sophisticated classification architectures, fine-tuning strategies, and introducing inductive biases to input via question-type conditioning. We also conducted further analyses and empirical comparisons, including detection of unimodal biases in training and evaluation data, fine-grained analyses across different question types, and comparisons across different multilingual Transformer models and transfer scenarios. We hope that this work will spark more interest and inspire future research on cross-lingual VQA tasks in particular, as well as on multilingual multimodal learning in general.
 
\section{Limitations}

Our study focuses on the cross-lingual VQA task relying on the xGQA dataset only. xGQA contains seven typologically different languages and many low-resource languages are not included. We aim to extend this study to other low-resource languages in the future, and to other datasets that were made publicly available after the completion of this study~\cite{Changpinyo:2022arxiv}.

In one part of our study, we assume gold question-type information is available during training and testing. This assumption is made for analysis purposes, in practice, one could train a classifier for question-type classification first.

The proposed self-bootstrapping method requires the ability to divide training into stages and reset weights during training.

We have averaged our results over four runs. From our experiments, we noticed that both of the underlying multilingual multimodal Transformers produced high variance results. We plan to investigate the causes of the variance in detail as part of future research. Currently, we relied on one established few-shot learning paradigm, recently~\citet{fabian} shows that combining English and target-language data might yield more robust transferring results, which we plan to investigate into future.

\section*{Acknowledgements}
{\scriptsize\euflag} 
The Ubiquitous Knowledge Processing Lab acknowledges the financial support of the German Federal Ministry of Education and Research (BMBF) under the promotional reference 13N15897 (MISRIK), and the LOEWE initiative (Hesse, Germany) within the emergenCITY center.
The work of Ivan Vuli\'{c} and Anna Korhonen has been supported by the ERC PoC Grant MultiConvAI: Enabling Multilingual Conversational AI (no. 957356) and a research donation from Huawei. The work of Ivan Vuli\'{c} was also supported in part by a personal Royal Society University Research Fellowship (no 221137; 2022-).

We thank Gregor Geigle for insightful feedback and suggestions on a draft of this paper.

\bibliographystyle{acl_natbib}
\bibliography{anthology,custom}

\clearpage 
\appendix

\section{Details of Training Setup and Hyperparameters}
\label{app:hpm}

The hyperparameters used to train \mmmp/ and \ucc/ models are summarized in Table~\ref{tab:app_hyp}. We conducted all experiments with either an NVIDIA V100 or A100 GPU. The numbers of training epochs across different model configurations are summarized in Table~\ref{tab:app_epochs}. Training time for the rest of our zero-shot experiments ranges from 8 to 24 hours. 

We searched over the following learning rates: 2e-5, 5e-5, and 1e-4.

We note that experiments which rely on Full GQA data ($-full$) have a significantly different training budget. This setup followed the previously recommended training setup of \citet{Li2020OscarOA}.

We use pretrained, state-of-the-art Transformer-based M3P and UC2 models (open-sourced), which build on pre-extracted image features from pretrained object detectors. M3P was pretrained via masked language modeling, cross-lingual masked language modeling and cross-modal text-image region alignments objectives. UC2 was trained similarly to M3P with an additional auxiliary task (i.e. translation). 

We extracted image features for M3P using the ResNet-101 backbone using the \texttt{vqa-maskrcnn-benchmark} model~\cite{massa2018mrcnn} (100 bounding boxes), and we extracted image features for UC2 using the bottom-up-attention~\cite{btan} (100 bounding boxes). The feature extraction procedures are different because the pretrained M3P and UC2 use different features. For experiments, we implemented everything in PyTorch, and we utilized Hugging Face Transformers~\cite{hftransformers} and MMT-Retrieval~\cite{mmretrieval}.

\begin{table}[h]
    \centering
    {\footnotesize
    \begin{tabularx}{\columnwidth}{l Y}
    \toprule
  \textbf{Name}  & Value \\
 \cmidrule(lr){1-1}  \cmidrule(lr){2-2}
  learning rate (\mmmp/) & 0.00002 \\
  learning rate (\ucc/) &  0.0001  \\
  train batch size & 192 \\
  warmup steps & 0 \\
  weight decay & 0.05 \\
  max grad norm & 1 \\
  dropout rate & 0.5 \\
  max seq length & 70 \\
  max img seq length & 50 \\
  $f_{trans}$ hidden dim & 768 \\
  optimizer & AdamW \\
    \bottomrule
    \end{tabularx}
    }%
    \caption{Hyperparameters.}
    \label{tab:app_hyp}
\end{table}

\begin{table}[h]
    \centering 
    \footnotesize
    \setlength{\tabcolsep}{2.5pt}
    \begin{tabular}{l cc cc}
    \toprule
    & Balanced & Balanced &  & \\
  \textbf{Exp.}  & Stage 1 & Stage 2 & Total Ep. & Time\\
 \cmidrule(lr){1-1}  \cmidrule(lr){2-3} \cmidrule(lr){4-5}
 \mdoq/  & 6 & - & 6  & $<$24hrs \\
 M3P$^{Q}$ + FT $_{short}$ & 4 & - & 4  & $<$24hrs\\
 M3P$^{Q}$ + FT $_{long}$ & 6 & - & 6  & $<$24hrs\\
 \msboot/ & 4 & 2 & 6 & $<$24hrs\\
 \cmidrule(lr){1-1}  \cmidrule(lr){2-3} \cmidrule(lr){4-5}
  \ucdoq/  & 6 & - & 6  & $<$24hrs \\
 UC2$^{Q}$ + FT $_{short}$ & 3 & - & 3  & $<$24hrs\\
 UC2$^{Q}$ + FT $_{long}$ & 6 & - & 6  & $<$24hrs \\
 \ucsboot/ & 3 & 3 & 6  & $<$24hrs\\
 \midrule
 
   & Full & Balanced &  & \\
  \textbf{Exp.}   & Stage 1 & Stage 2 & Total Ep. & Time\\
 \cmidrule(lr){1-1}  \cmidrule(lr){2-3} \cmidrule(lr){4-5}
  $-full$ & 5 & 2 & 7  & 4 days\\
    \bottomrule
    \end{tabular}
    \caption{Training epochs and times. Full and Balanced indicate the GQA subset used for training. The self-bootstrapping experiments are initialized from the weights of $short$ experiments.}
    \label{tab:app_epochs}
\end{table}

\begin{table}[t!]
    \centering
    {\footnotesize
    \begin{tabularx}{\columnwidth}{l Y}
    \toprule
    \textbf{Question Type} & Count \\
 \cmidrule(lr){1-1}  \cmidrule(lr){2-2}
        Verify &  2,251 \\
        Logical &  1,803 \\
        Compare &  5,89 \\
        Query &  6,804 \\
        Choose &  1,129 \\
    \bottomrule
    \end{tabularx}
    }%
    \caption{GQA test-dev set: distribution of questions over question types.}
    \label{tab:app_dist}
\end{table}
\begin{table}
\footnotesize
    \centering  
    \def\arraystretch{0.9}
    \setlength{\tabcolsep}{2.6pt}
    \begin{tabular}{l cccc}
    \toprule
    \textbf{Question Type} & \mmmp/ & \mmmp/ + SB & \mdoq/ & \msboot/ \\
   \cmidrule(lr){1-1} \cmidrule(lr){2-5}
    Verify	& 40.15 & 44.45& \textbf{58.35} & 55.98 \\
    Logical	& 39.15 & 45.29& \textbf{53.89} & 53.65 \\
    Compare	& 35.95 & 40.75& 45.82 & \textbf{47.14} \\
    Query & \textbf{24.57} & 21.42& 21.86 & 24.50 \\
    Choose & 30.63 & 29.07& 29.43 & \textbf{33.57} \\
    \midrule
    \textbf{Question Type} & \ucc/ & \ucc/ + SB & \ucdoq/ & \ucsboot/ \\
    \cmidrule(lr){1-1} \cmidrule(lr){2-5}
    Verify & 41.55 & 51.27 & 59.94 & \textbf{61.70} \\
    Logical& 35.40 & 48.32 & 54.87 & \textbf{56.49}  \\
    Compare	& 34.48 & 44.71 & 49.15 & \textbf{51.73}  \\
    Query & 23.18 & 27.68 & 23.94 &  \textbf{27.88} \\
    Choose & 29.51 &36.62 & 29.66 &  \textbf{36.14} \\
    \bottomrule
    \end{tabular}
    \caption{Average accuracy on different structural question types from xGQA (excluding English). \mmmp/ and \ucc/ are using Deep architecture.}
    \label{tab:struct}
\end{table}

\section{Structural Question Types in GQA and xGQA}
\label{app:gqa_qtype}
There are 5 different structural question-types in GQA and, consequently, in xGQA. We used the exact lowercase name of each question type as the \texttt{QType} token in our experiments, namely: \textit{verify, logical, compare, query, and choose}. The text input follows the format of: \texttt{`[QType] : [Question]'} (see again \S\ref{sec:qtype}). Some example questions for each question type are as follows:

\vspace{0.2em}
\noindent\textbf{Verify:} Yes/No questions. E.g.\textit{ Do you see books near the device that looks gray? Is the bus blue?}

\vspace{0.2em}
\noindent\textbf{Logical:} Questions that require logical inference. E.g. \textit{Is there any motorcycle or ball in the scene? Does the dirt look brown and fine?}

\vspace{0.2em}
\noindent\textbf{Compare:} Comparison questions between two or more objects. E.g. \textit{Who seems to be younger, the boy or the woman?}

\vspace{0.2em}
\noindent\textbf{Query:} Open questions. E.g. \textit{What color are the pants? What is the animal that is standing on the grass called?}

\vspace{0.2em}
\noindent\textbf{Choose:} Choose from two presented alternatives. E.g. \textit{Is it red or blue? What size is the jacket, small or large?}

Verify and Logical question types are binary question types (Yes/No). The question type distribution in the test-dev set of GQA is given in Table~\ref{tab:app_dist}, while we provide average accuracy scores overall target languages in xGQA (excluding English), with a representative set of models, in Table~\ref{tab:struct}.

\renewcommand{\arraystretch}{0.7}
\begin{table*}[h]
    \setlength{\tabcolsep}{3pt}
    \centering
    \footnotesize
    \begin{tabular}{l c ccccccc c}
    \toprule
    \textbf{Method} & En & De & Zh & Ko & Id & Bn & Pt & Ru & \textbf{Avg}\\
    \cmidrule(lr){1-1} \cmidrule(lr){2-2} \cmidrule(lr){3-10} 
     \mmmp/ (Linear) & 
        51.88$\scriptscriptstyle\pm0.7$ &
        27.45$\scriptscriptstyle\pm5.8$ &
        16.33$\scriptscriptstyle\pm8.3$ &
        13.70$\scriptscriptstyle\pm5.4$ &
        25.25$\scriptscriptstyle\pm11.4$ &
        10.59$\scriptscriptstyle\pm3.4$ &
        21.10$\scriptscriptstyle\pm3.4$ &
        20.95$\scriptscriptstyle\pm3.3$ &
        19.34
        \\
    \mmmp / w/ LN & 
        51.66$\scriptscriptstyle\pm0.6$ &
        35.33$\scriptscriptstyle\pm5.4$ &
        27.80$\scriptscriptstyle\pm10.9$ &
        25.55$\scriptscriptstyle\pm11.4$ &
        30.54$\scriptscriptstyle\pm9.8$ &
        17.94$\scriptscriptstyle\pm8.6$ &
        30.61$\scriptscriptstyle\pm7.2$ &
        29.74$\scriptscriptstyle\pm6.6$ &
        28.22 \\
    \mmmp / w/o LN & 
    50.89$\scriptscriptstyle\pm1.0$ &
    32.92$\scriptscriptstyle\pm5.6$ &
    22.14$\scriptscriptstyle\pm8.0$ &
    20.33$\scriptscriptstyle\pm9.1$ &
    25.44$\scriptscriptstyle\pm6.5$ &
    16.88$\scriptscriptstyle\pm8.0$ &
    29.40$\scriptscriptstyle\pm7.8$ &
    29.31$\scriptscriptstyle\pm7.9$ &
    25.20 \\
    \midrule
    \ucc/ (Linear) & 
        57.83$\scriptscriptstyle\pm	0.3$ &
        40.57$\scriptscriptstyle\pm	1.7$ &
        35.54$\scriptscriptstyle\pm	3.4$ &
        16.95$\scriptscriptstyle\pm	6.1$ &
        34.18$\scriptscriptstyle\pm	0.8$ &
        8.53$\scriptscriptstyle\pm	1.9$ &
        24.90$\scriptscriptstyle\pm	3.7$ &
        24.05$\scriptscriptstyle\pm	4.6$ &
        26.39	 
        \\
    \ucc/ w/ LN &  
        58.31$\scriptscriptstyle\pm0.2$ &
        41.33$\scriptscriptstyle\pm1.6$ &
        34.77$\scriptscriptstyle\pm2.2$ &
        23.87$\scriptscriptstyle\pm1.5$ &
        34.79$\scriptscriptstyle\pm1.3$ &
        11.82$\scriptscriptstyle\pm1.9$ &
        29.30$\scriptscriptstyle\pm4.5$ &
        29.41$\scriptscriptstyle\pm3.7$ &
        29.33 \\
    \ucc/ w/o LN & 
58.03$\scriptscriptstyle\pm0.5$ &
42.74$\scriptscriptstyle\pm1.4$ &
37.84$\scriptscriptstyle\pm3.0$ &
24.91$\scriptscriptstyle\pm5.2$ &
33.56$\scriptscriptstyle\pm1.6$ &
13.21$\scriptscriptstyle\pm4.5$ &
29.99$\scriptscriptstyle\pm4.5$ &
29.47$\scriptscriptstyle\pm6.3$ &
30.25\\
    \bottomrule
    \end{tabular}
    \caption{Zero-shot cross-lingual transfer results with and without LayerNorm.}
    \label{app:tab_ln}
\end{table*}

\begin{table*}[h]
    \setlength{\tabcolsep}{3pt}
    \centering
    \footnotesize
  \begin{tabular}{l c ccccccc c}
    \toprule
    \textbf{Method} & En & De & Zh & Ko & Id & Bn & Pt & Ru & \textbf{Avg}\\
    \cmidrule(lr){1-1} \cmidrule(lr){2-2} \cmidrule(lr){3-10} 
    \ucsboot/ - full & 
    57.88$\scriptscriptstyle\pm0.2$& 
    50.52$\scriptscriptstyle\pm0.5$& 
    47.63$\scriptscriptstyle\pm0.2$& 
    37.56$\scriptscriptstyle\pm1.7$& 
    40.37$\scriptscriptstyle\pm1.6$& 
    25.25$\scriptscriptstyle\pm1.4$& 
    40.56$\scriptscriptstyle\pm0.2$& 
    41.67$\scriptscriptstyle\pm0.8$& 
    40.51  \\
    \bottomrule
    \end{tabular}
    \caption{Zero-shot results when the models are trained with Full GQA data.}
    \label{tab:main_full}
\end{table*}

\section{Classification Architecture with and without Layer Normalization}
\label{app:d}

\begin{figure}[]
  \centering
   \includegraphics[scale=0.6]{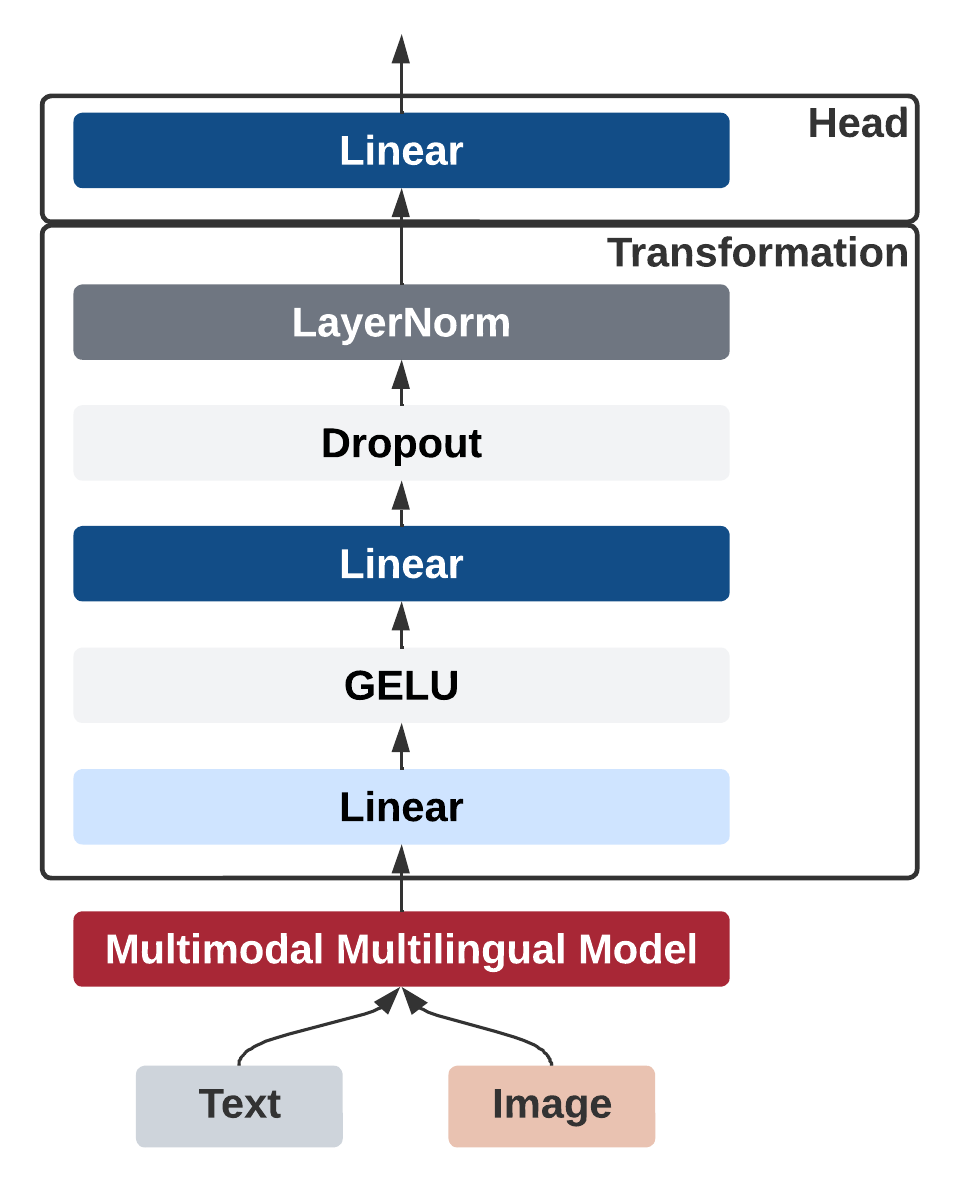}
    \caption{The deep(er) classification architecture (see \S\ref{ss:prediction}). The first linear layer in the transformation uses an orthogonal initializer.}%
     \label{fig:arch}
\end{figure}

The deeper variant of the classification architecture from \S\ref{ss:prediction} is illustrated in Figure~\ref{fig:arch}. The \textit{Multimodal Multilingual Model} block in Figure~\ref{fig:arch} denotes one of the two pretrained multimodal multilingual models used throughout the (main) paper: UC2 and M3P.

We further experimented with another variant of the architecture, where we removed the layer normalization (LayerNorm) layer. The results of this variant are available in Table~\ref{app:tab_ln}.

In a nutshell, LayerNorm has more impact on M3P's zero-shot transfer accuracy scores than on UC2. However, the variance of UC2 results increases with the removal of LayerNorm.

\section{Accuracy vs.~Total Training Epochs}
\label{app:acc_vs_time}

We conducted experiments with different total numbers of training epochs with \mmmp/to understand the effect of the self-bootstrapping fine-tuning strategy. We experimented with the following three model configurations across different setups:
\begin{enumerate}
    \item M3P$^{Q}$ + FT: We train the \mdoq/ model with text embeddings frozen for 4, 6 and 10 epochs.
    \item M3P$^{Q*}$ + FT: We initialize the \mdoq/ model with fine-tuned weights (including transformation, classification head) from 1 (i.e., the variant above), and train for 4 epochs. We continue to fine-tune the model for 2 or 5 more epochs after resetting the learning rate and the optimizer.  
    \item \msboot/: We train the \mdoq/ model with self-bootstrapping and the classification head weights from variant 1 above and do it for 4 epochs. We continue to fine-tune the model for 2 or 5 epochs.
\end{enumerate}

\begin{figure}[t!]
    \centering
    \begin{subfigure}[t!]{0.45\textwidth}
        \includegraphics[width=\linewidth]{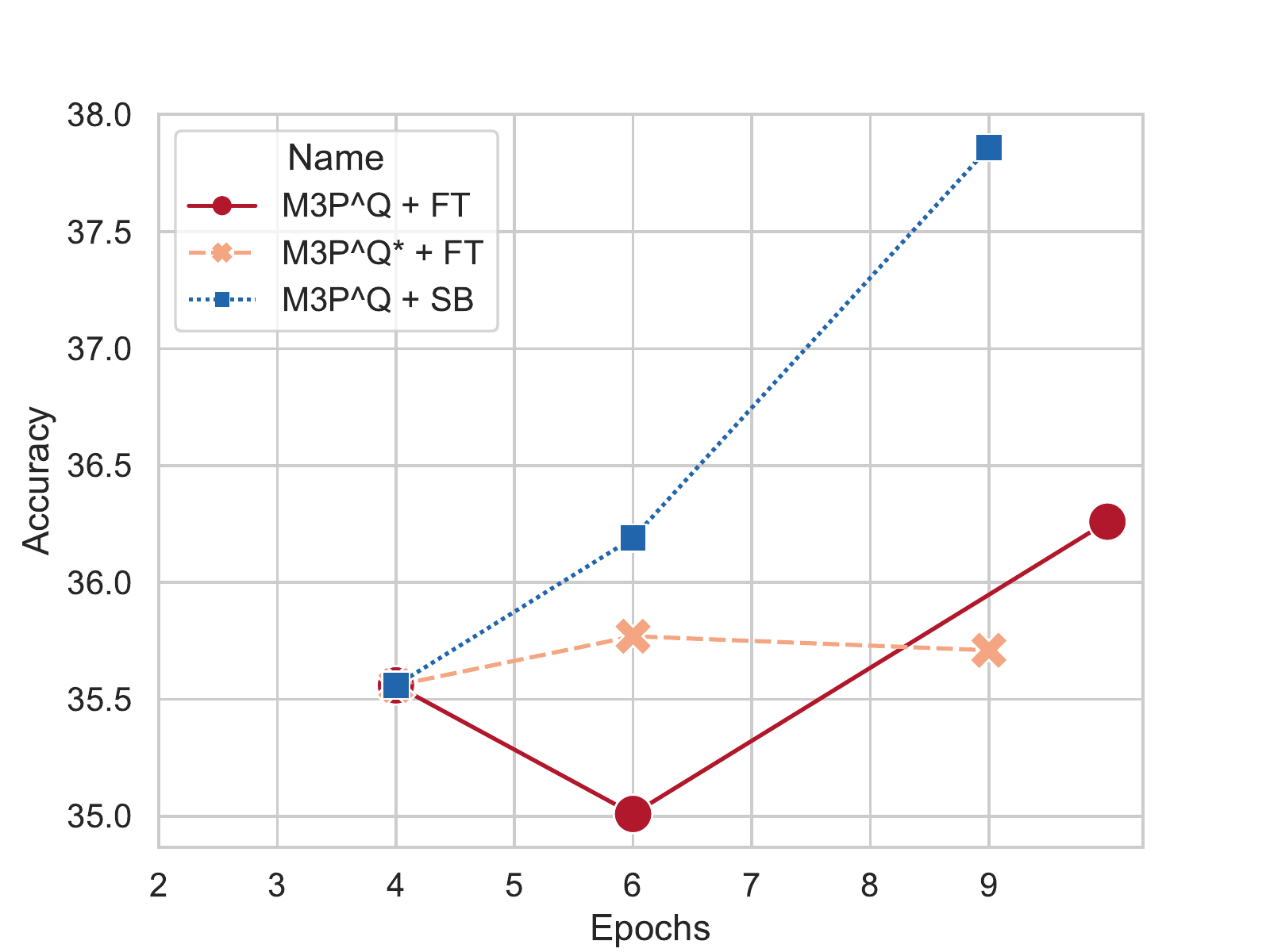}
        \caption{M3P}
        \label{fig:m3p_acc_vs_epoch}
    \end{subfigure}
    \begin{subfigure}[t!]{0.45\textwidth}
        \includegraphics[width=\linewidth]{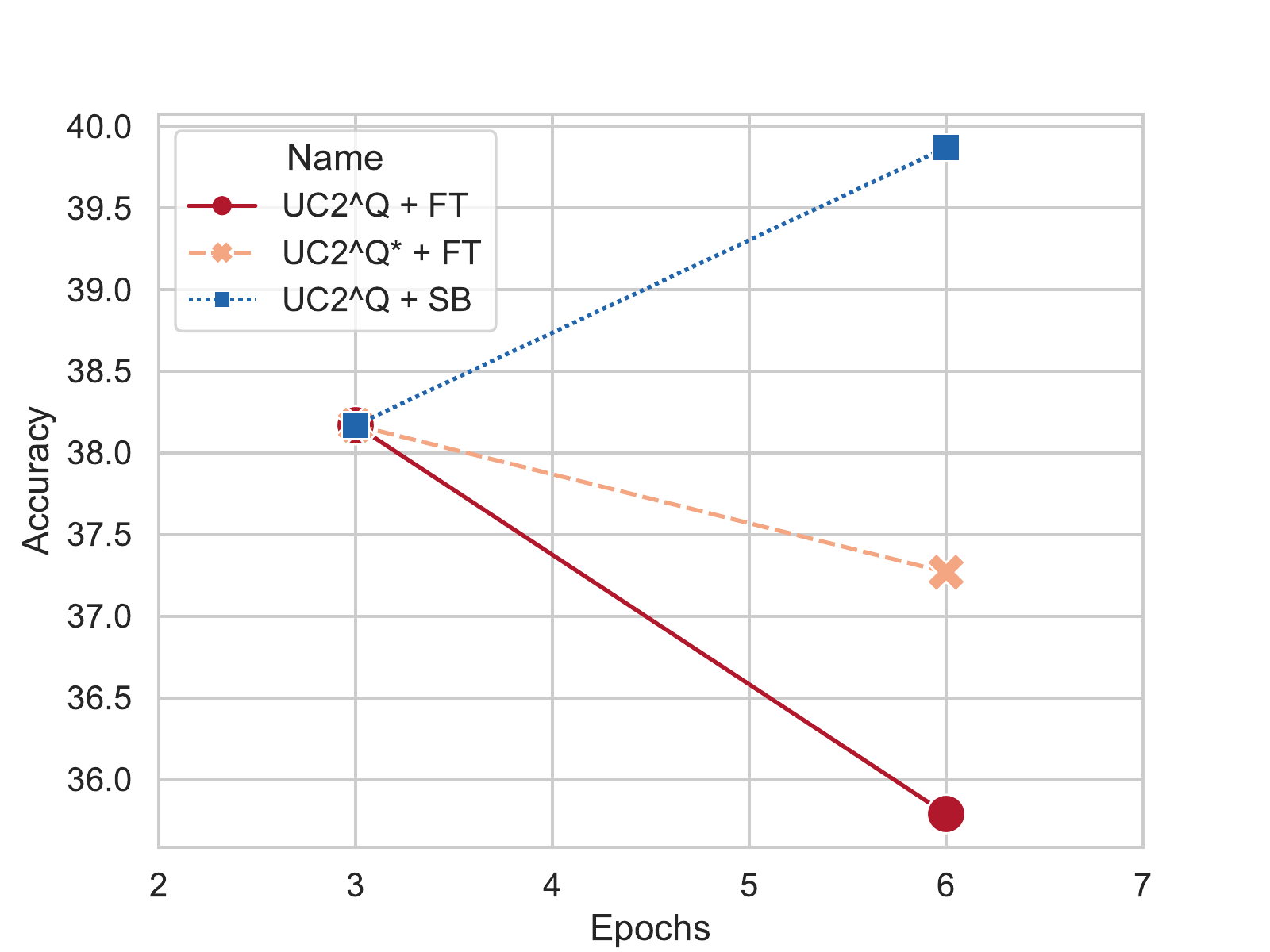}
        \caption{UC2}
        \label{fig:uc2_acc_vs_epoch}
    \end{subfigure}
    \vspace{-1.0mm}
    \caption{Average accuracy versus total training epochs.}
    \label{fig:epochs_appendix}
    \vspace{-1.5mm}
\end{figure}

We also run similar variants with UC2 as the underlying model with shorter training epochs. These variants are UC2$^{Q}$ + FT / UC2$^{Q*}$ + FT / \ucsboot/ where superscripts and acronyms remain the same as the M3P variants.  Results of these experiments are provided in Figure~\ref{fig:m3p_acc_vs_epoch} (M3P) and Figure~\ref{fig:uc2_acc_vs_epoch} (UC2).

We observe that the gains in cross-lingual transfer with +FT variants diminish or even start decreasing with the increase of training time. Similar results are observed when we reset the learning rate, weight decay and optimizer after training for 4 epochs. We also find that self-bootstrapping training continually improves the results, even with less additional total training epochs.

Moreover, the performance of self-bootstrapping is considerably more stable (lower variance) across random seeds, even though its classification heads are initialized from the corresponding trained weights from the M3P$^{Q}$ + FT experiments.

We also observe an increase in zero-shot transfer accuracy scores with more epochs of training in Stage 2 of self-bootstrapping. However, this results in much longer training times, which may not be realistic for academic and even some industry settings.

\section{Results with Full GQA Data}
\label{app:full}
It is worth noting that the experiments trained with full GQA data ($-full$) have a significantly different (and larger) training budget (see \S\ref{sec:further_exp}). We follow the previously recommended total training budget of \citet{Li2020OscarOA} and combine it with our self-bootstrapping fine-tuning strategy. Table~\ref{tab:main_full} shows the detailed results.

\section{Few-shot Experiments: Full Results}
\label{app:few_shot}

Table~\ref{tab:app_few_shot} shows the detailed results of our few-shot experiments, where the summary table is provided in the main paper: Table~\ref{tab:fs} in \S\ref{sec:further_exp}.

\renewcommand{\arraystretch}{0.6}
\begin{table}[h]
    \footnotesize
    \centering
    \begin{tabular}{ll cccc}
\toprule
        Lang & \textbf{Method} & 0 & 1 & 5 & 48  \\
\cmidrule(lr){2-2} \cmidrule(lr){3-3} \cmidrule(lr){4-6}
    de  & \mmmp/&  39.45&	40.76&	41.88&	44.20 \\
    & \mmmp/ + SB & 39.25&	39.72&	40.98& 43.08 \\
    & \mdoq/ & 36.84& 	38.28& 	40.11& 		43.18\\
    & \msboot/ & 40.99&	40.74&	40.39&		41.71\\
\cmidrule(lr){2-2} \cmidrule(lr){3-3} \cmidrule(lr){4-6}
    zh & \mmmp/ &35.76&	37.65&	40.28&		42.18
\\
    & \mmmp/ + SB & 32.96&	35.55	&38.24	&	41.15 \\
    & \mdoq/ &33.74& 	35.97& 	37.95	& 	41.28
\\
    & \msboot/ &36.95&	36.88&	37.60&		39.38
\\
\cmidrule(lr){2-2} \cmidrule(lr){3-3} \cmidrule(lr){4-6}
    ko  & \mmmp/ & 34.53&	36.58&	36.79&		39.61
\\
    & \mmmp/ + SB & 36.04&	36.92&	37.31&		39.41 \\
    & \mdoq/ &31.96& 	32.77& 	35.39	& 	40.45
 \\
     & \msboot/ &35.78&	35.38&	37.46&		38.99
\\
\cmidrule(lr){2-2} \cmidrule(lr){3-3} \cmidrule(lr){4-6}
    id & \mmmp/ &38.38&	39.39&	40.63&		42.57
 \\
    & \mmmp/ + SB & 29.17&	36.94& 39.49 &	41.16\\
    & \mdoq/ & 34.69& 	35.37& 	38.50	& 	42.12
\\
     & \msboot/ & 37.75&	36.25&	38.57&		39.92
\\
\cmidrule(lr){2-2} \cmidrule(lr){3-3} \cmidrule(lr){4-6}
    bn & \mmmp/ &  24.27&	30.53&	34.72&		40.73
\\
    & \mmmp/ + SB & 22.71&	25.94&	33.96&	40.46 \\
    & \mdoq/ & 27.67& 	29.95& 	33.15	& 	40.36\\
     & \msboot/ & 30.50&	31.77&	34.08&		39.24
\\
\cmidrule(lr){2-2} \cmidrule(lr){3-3} \cmidrule(lr){4-6}
    pt & \mmmp/ & 38.19&	38.35&	40.54&		44.27
\\
    & \mmmp/ + SB & 38.17&	37.98&	39.35&		43.01\\
    & \mdoq/ & 36.87& 	37.93& 	39.72	& 	43.08
\\
     & \msboot/ & 38.56&	39.24&	39.71&		40.56
\\
\cmidrule(lr){2-2} \cmidrule(lr){3-3} \cmidrule(lr){4-6}
     ru & \mmmp/ & 38.46&	40.06&	40.22&		42.38\\
     & \mmmp/ + SB & 37.84 &	38.20&	38.54&		41.95\\
     & \mdoq/ &34.86& 	37.51& 	39.82	& 	42.64\\ 
     & \msboot/ & 38.76&	39.74&	39.39	&	40.19\\
    \midrule
    de & \ucc/ & 40.39 &	44.23 &	46.03 &	49.51\\
    &\ucc/ + SB & 49.52&	50.10&	50.30&	51.42\\
    & \ucdoq/ & 46.26&	46.95&	46.94&	49.42\\
    & \ucsboot/ & 50.23&	50.70&	50.53&	51.39\\
\cmidrule(lr){2-2} \cmidrule(lr){3-3} \cmidrule(lr){4-6}
    zh & \ucc/ & 37.26 &	41.70 &	42.68 &	46.32
\\
   & \ucc/ + SB &43.54&	46.30&	47.17&	48.80\\
    & \ucdoq/  & 43.89&	44.90&	45.56&	47.24
 \\
    & \ucsboot/& 46.37&	47.82&	48.32&	48.47
\\
\cmidrule(lr){2-2} \cmidrule(lr){3-3} \cmidrule(lr){4-6}
    ko & \ucc/  & 25.93 &	32.63 &	36.11 &	41.11
  \\
  &\ucc/ + SB & 36.48&	36.73&	37.84&	43.90\\
    & \ucdoq/ &32.45&	35.79&	37.37&	42.04
\\
    & \ucsboot/ & 37.80&	39.05&	40.68&	43.38
\\
\cmidrule(lr){2-2} \cmidrule(lr){3-3} \cmidrule(lr){4-6}
    id & \ucc/ & 35.76 &	39.35 &	40.12 &	44.24
\\
    &\ucc/ + SB & 32.70&	38.18&	42.88&	47.06\\
     & \ucdoq/ & 36.70&	39.54&	41.40&	45.78
 \\
    & \ucsboot/ & 38.34&	42.16&	42.33&	47.01
\\
\cmidrule(lr){2-2} \cmidrule(lr){3-3} \cmidrule(lr){4-6}
    bn & \ucc/ &12.00 &	21.91 &	25.95 &	39.75
 \\
 &\ucc/ + SB & 24.66&	29.76&	32.31&	42.08\\
    & \ucdoq/&  25.29&	27.68&	32.75&	39.82
\\
    & \ucsboot/& 24.07&	31.67&	35.77&	42.83
\\
\cmidrule(lr){2-2} \cmidrule(lr){3-3} \cmidrule(lr){4-6}
    pt & \ucc/ &29.79 &	33.86 &	40.18 &	45.23
 \\
 &\ucc/ + SB & 38.79&	40.49&	41.95&	47.34\\
    & \ucdoq/ & 36.60&	39.56&	40.67&	46.45
\\
    & \ucsboot/& 40.36&	42.65&	43.79&	47.63
\\
\cmidrule(lr){2-2} \cmidrule(lr){3-3} \cmidrule(lr){4-6}
    ru & \ucc/ & 29.94 &	38.97 &	39.66 &	44.41
\\
    &\ucc/ + SB & 40.93&	42.02&	42.54&	46.15\\
    & \ucdoq/ & 39.76&	40.26&	41.46&	45.04
\\
    & \ucsboot/ &41.62&	42.42&	44.32&	45.63
 \\
     \bottomrule
    \end{tabular}
    \caption{Few-shot transfer average accuracy with different amounts of training data. \mmmp/ and \ucc/ are using the deeper classification architecture.}
    \label{tab:app_few_shot}
\end{table}

\end{document}